\documentclass{article}

\usepackage{arxiv}
\usepackage{natbib}
\usepackage[utf8]{inputenc} 
\usepackage[T1]{fontenc}    
\usepackage{hyperref}       
\usepackage{url}            
\usepackage{booktabs}       
\usepackage{amsfonts}       
\usepackage{nicefrac}       
\usepackage{microtype}      
\usepackage{lipsum}
\usepackage{graphicx}
\usepackage{algorithm}
\usepackage{algpseudocode}
\usepackage{amsmath,bm}
\graphicspath{ {./images/} }

\title{Functional Subspace, where language models can use vector algebra to solve problems}

\author{%
	Jung H. Lee \\
	Pacific Northwest National Laboratory\\
	Seattle, WA \\
	\texttt{jung.lee@pnnl.gov} \\
	\And
	Sujith Vijayan \\
	School of Neuroscience\\
	Virginia Tech\\
	Blacksburg, VA \\
	\texttt{neuron99@vt.edu} \\
}

\begin{document}
\maketitle
\begin{abstract}
	Large language models (LLMs) were invented for natural language tasks such as translation, but they
	have proved that they can perform highly complex functions across domains. Additionally, they have
	been thought to develop new skills without being trained on them. These learning capabilities lead to  LLMs adoption in a wide range of domains. Thus, it is imperative that we understand their operating mechanisms and limitations for proper diagnostics and repair. The earlier studies proposed that high level concepts are encoded as linear directions in LLMs activation space and that the geometry of embeddings have semantic meanings. Inspired by these studies, we hypothesize that LLMs may use subspaces and vector algebra in subspaces to perform tasks. To address this hypothesis, we analyze LLMs' functional modules and residual streams collected from LLMs engaging in in-context learning (ICL), one of the emergent abilities. Our analyses suggest that 1) LLMs can create subspaces, where evidence can be accumulated and 2) ICL tasks can be solved via simple algebraic operations in subspaces. 
\end{abstract}

\section{Introduction}
The explosive growth in LLMs applications indicates that LLMs are highly capable learners \cite{10.1145/3744746, laskar-etal-2024-systematic, minaee2025largelanguagemodelssurvey}. After being pretrained to predict the next tokens, LLMs can be fine-tuned for complex tasks. Notably, a line of studies suggests that fine-tuning may not be strictly necessary, as LLMs can obtain “emergent” abilities \cite{matarazzo2025, lu-etal-2024-emergent, icl1, icl2}. Pretraining does not aim to train LLMs to write code, but still, after pretraining, they can generate code. However, it remains poorly understood how LLMs can generalize and develop emergent abilities, making the diagnosis of LLMs’ operations almost impossible. Before LLMs are being fully utilized in safety-critical domains, it is imperative to better understand the principles behind LLMs’ generalization and emergent abilities to diagnose and repair LLMs’ operations

To this end, we investigate how LLMs support in-context learning (ICL), one of LLMs’ emergent abilities \cite{icl1, icl2}. ICL allows users to reconfigure LLMs to perform specific tasks without retraining or fine-tuning. For instance, a pretrained LLM can function as a thesaurus to an input prompt query, if the input prompt provides a pair of synonyms. ICL tasks are often simpler than general tasks such as coding, but LLMs can still 1) extract task information from input prompts and 2) apply the information to answer users’ queries. These two basic functions are essential for LLMs to interpret users’ intentions and generate correct responses, which means that investigating ICL may lead us to a better understanding of how LLMs perform more general tasks. 

Our study is inspired by two lines of studies. First, linear representation hypothesis posits that high level concepts are encoded as linear directions in LLMs’ activation space \cite{park2023the, pmlr-v235-jiang24d}. According to this hypothesis, tokens (i.e., sub-words) associated with `happiness’ are mapped onto vectors pointing to a `happy’ direction, and those associated with `unhappiness’ are mapped onto vectors pointing to a `sad’ direction. Second, the geometry of word embeddings may be linked to semantic meanings. For example, in the embedding space, the word `king' can be converted to the word `queen' if it is subtracted by `man' and added by `woman' \cite{mikolov-etal-2013-linguistic, wordembedding}.

Based on these studies, we hypothesize that LLMs can support ICL by projecting input tokens into a subspace, where a desired task can be solved by using vector algebra. To address this hypothesis, we use principal component analysis (PCA) and linear regression analysis to determine the existence of a LLM subspace, where the answer can be inferred by vector algebra. Our empirical evaluations suggest that ICL tasks can be considered vector algebra problems in some of the LLMs' subspaces. 

\section{Subspace, naturally generated by LLMs}
In this section, we discuss how transformers' functional modules generate subspaces and use them to support ICL. In section \ref{icl}, we briefly discuss ICL tasks used in this study, and in section \ref{mechanism1}, we discuss potential mechanisms by which LLMs' functional modules may use subspaces to support ICL. 
\subsection{ICL tasks}\label{icl}

\begin{figure}
	\begin{center}
		\includegraphics[width=1\linewidth]{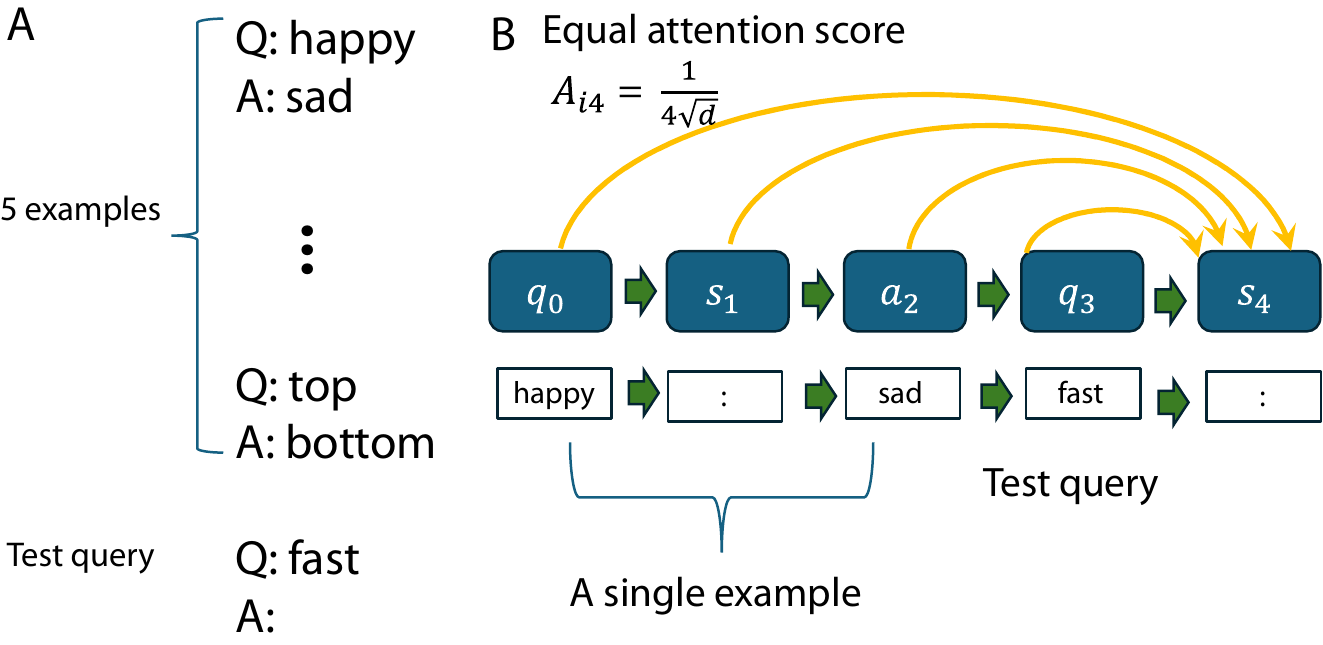}
	\end{center}
	\caption{ICL prompts. (A), The structure of the used ICL prompts. (B), The structure of a single example ICL with uniform attention score $A_{ij}$. This panel illustrates how symbols $q_i$, $s_i$, $a_i$ are mapped onto ICL tokens.}
	\label{fig:diagram}
\end{figure}

\citet{icl1} found that LLMs can infer correct answers, if they are exposed to a few examples of desired tasks; see \cite{icl2} as well. When a few pairs of synonyms are included in the input prompts, LLMs return a synonym to the last query word. Thus, ICL prompts contain multiple examples of desired tasks and a query.  In our study, we use the dataset generator released by an earlier study \cite{todd2024function} to create ICL prompts for $6$ different tasks, antonyms, synonyms, countries-their capital cities (country-capital),  English-French words, company-product and person-sport. All $6$ tasks can be demonstrated with two words separated by a separator. Throughout this study, we refer to them as queries, separators and answers. For all tasks, we provide $5$ example pairs of queries and answers and a single test query (without a corresponding answer) and direct LLMs to find its counterpart using example pairs. Each query and answer are marked by the separators, `Q:' and `A:' respectively (Fig. \ref{fig:diagram}A). 

\subsection{Transformer blocks and subspaces}\label{mechanism1}
LLMs first convert words in input prompts into tokens (i.e., sub-words) and use a sequence of transformer layers to process them progressively, each of which consists of self-attention (SA) layer and Feed-Forward Network (FFN) \cite{vaswani2023attention}. A transformer layer $l$ receives inputs $h^{l-1}$ from its previous layer and generates $h^l$ using its own SA and FFN. As the transfer layer uses residual connections, $h^l$ can be recursively expressed as $h^l=h^{l-1}+a^{l}+m^{l}(a^l+h^{l-1})$ \cite{meng2023massediting}, where $a^l$ and $m^l$ denote the output of SA layer and FFN in transformer layer $l$, respectively. That is, a transformer block $l$ combines a token $h^{l-1}_i$ from a previous layer with the output of SA layer $a^l_i$. It is notable that $a^l_i$ allows a token $h^{l-1}_i$ to understand the context in the preceding texts. 

\subsubsection{FFNs}\label{mechanism2}
Recent studies have suggested that FFNs can function as associative memory modules that can accept a key and return a value\cite{key-value, meng2022locating, meng2023massediting}. For instance, when a FFN accepts the ‘Space Needle’ as an input (i.e., a key), it returns `Seattle' as an output (i.e., a value) \cite{meng2022locating}. We note that human words often convey multiple meanings in different contexts, suggesting that FFNs need to store multiple values for a single key and choose the right one accordingly. For instance, if a LLM is asked to explain the function of the Space Needle, it should recall that it is an observatory and retrieve the ‘observatory’ instead of ‘Seattle’. 

Then, how does a FFN store multiple keys and choose the right one according to the context evaluated by a SA layer? FFNs in LLMs consist of two synaptic layers, ($\bm{w}^{1st}$ and $\bm{w}^{2nd}$), and one hidden layer of hidden cells $m_i$ (Eq. \ref{eq-ffn}). As hidden cells $m_i$s operate independently, we consider the input $h^{l-1}_i+a^l_i$ and output $v^l_i$ of a single neuron $m_i$  
\begin{equation}\label{eq-ffn}
	\begin{split}
		x_i= w^{i}_{F} \cdot (h^{l-1}_j+a^l_i),\\
		v^l_i=w_Sg(x_i),
	\end{split}
\end{equation}

where $w_F$ and $w_S$ denote a row of $\bm{w}^{1st}$ and a column of $\bm{w}^{2nd}$, respectively, and $g$ is the activation function, which is commonly the ReLU or one of its variants. From this, it is clear that $m_i$ becomes active only when its input is aligned with $w^i_F$. \citet{key-value} suggested that hidden neurons $m_i$ in a FFN were indeed memory cells. In this case, $w^i_F$ can be considered a feature. More importantly, this observation raises the possibility that a FFN can encode semantic meanings of human words with lower concepts (i.e., features). 

In the example above, three memory cells may be sufficient for a FFN to selectively recall Seattle or the observatory. Let's consider a case, in which three memory cells $m_0$, $m_1$ and $m_2$ of a FFN are configured to detect `Space Needle’, `geography’ and `facility’s function’.  When $m_0$ and $m_1$ are activated together by $h^{l-1}_i+a^l_i$, its output will be determined by the outputs of $m_0$ and $m_1$ and $v^l=\alpha w^0_S+\beta w^1_S$. When $m_0$ and $m_2$ are activated together, its output will be determined by the outputs of $m_0$ and $m_2$ and $\tilde{v}^l=\alpha w^0_S+\gamma w^2_S$. If a LLM can map $v^l$ and $\tilde{v}^l$ onto the token of Seattle and that of the observatory, it can selectively recall either Seattle or the observatory. 

As the Space Needle can be defined as its function or location, most words, if not all, can be defined with lower-level concepts. For example, a motor vehicle can be described with multiple features such as the number of its wheels and the inclusion of a cargo bed. This line of reasoning leads us to hypothesize that LLMs may be trained to use a large and unknown set of context vectors $c_k$s (e.g., geography or the facility's function in the example above) to encode semantic meanings of human words that are hierarchically organized, and these contexts can be stored and detected by FFNs. 

According to this hypothesis, single entities/words can be encoded by a finite number of context vectors $c_k$s. In other words, they are encoded in subspaces spanned by $c_k$s. 

\subsubsection{ICL as a vector operation} \label{mechanism3}
In this study, we treat an ICL task as a function that accepts a query token ($\vec{q}$) and a separator token ($\vec{s}$) and returns an answer token ($\vec{a}$) (Eq. \ref{eq:func}).

\begin{equation}\label{eq:func}
	\vec{a}=f(\vec{q}+\vec{s})
\end{equation}

Inspired by the observation that the embedding of the word ‘king’ is converted to that of the word ‘queen’ by linear operations in an embedding space, we hypothesize that the linear operations can also implement a necessary function $f$ in a subspace (Eq. \ref{eq_4}).

\begin{equation}\label{eq_4}
	\vec{a}=\alpha \vec{q}+\beta \vec{s} +\gamma
\end{equation}, where $\gamma$ is an intercept (i.e.,  constant). 

To address this hypothesis, we conduct the linear regression analysis with $30$ principal components using scikit-learn \cite{scikit-learn}. If $\vec{a}$ is well described by linear function of $\vec{q}$ and $\vec{s}$, the quality of linear regression would be high. Otherwise, its quality would be low. Below, we examine $R^2$ to evaluate the quality of the linear regression analysis with $30$ principal components (i.e., the potential bases for LLMs’ subspaces).

\section{Empirical evaluation}

In this study, we probe $6$ pretrained LLMs publicly available: GPT-j-6B \cite{gpt-j}, Meta-Llama-3.1-8B\cite{llama3}, OLMO-2-$0325$-32B\cite{olmo20242olmo2furious},  Phythia-12B\cite{pythia}, gemma-3-27b-it \cite{gemmateam2025gemma3technicalreport} and GPT-NEOX-$20$B\cite{gpt-neox}. All models are instantiated and tested with the publicly available machine learning libraries, Pytorch \cite{Paszke2017} and Transformers \cite{wolf2020huggingfacestransformersstateoftheartnatural}. All experiments and analyses are conducted using two consumer grade workstations. One workstation is equipped with Intel’s Core I9 CPU (64GB DRAM) and NVIDIA’s GTX 4090 (24GB VRAM), and the other, with AMD’s Strix Halo with 128GB unified memory. 

\subsection{Principal components as potential bases for subspaces}

Since residual streams of LLMs store all hidden representations, we analyze them to identify subspaces (i.e., components). Specifically, we create $200$ ICL prompts for $6$ different tasks  \cite{todd2024function}, antonyms, synonyms, countries-their capital cities (country-capital), English-French words, product-company and person-sport. Each prompt contains 5 examples and 1 question, and queries and answers are separated by separators (`:'); see Fig. \ref{fig:diagram}A. In the experiment, we collect residual streams of all tokens from all layers, while LLMs process ICL prompts. After collecting residual streams, we convert them into $2D$ arrays. Each row of the arrays is a residual stream, and thus the size of the row vectors is the same as the model dimension $d$. For instance, the model dimension $d$ of Meta-LLama-3.1-8B is $4096$. The number of rows is the same as the total number of tokens. As some words are tokenized into multiple tokens, the exact number of tokens vary from one model to another. As the residual streams are not bounded, each row is normalized to have a unit norm $\vec{s} \leftarrow N(\vec{s})$. Additionally, all columns are centered $\vec{s} \leftarrow S(\vec{s})$, as it is required for PCA. 

\begin{figure}
	\begin{center}
		\includegraphics[width=1\linewidth]{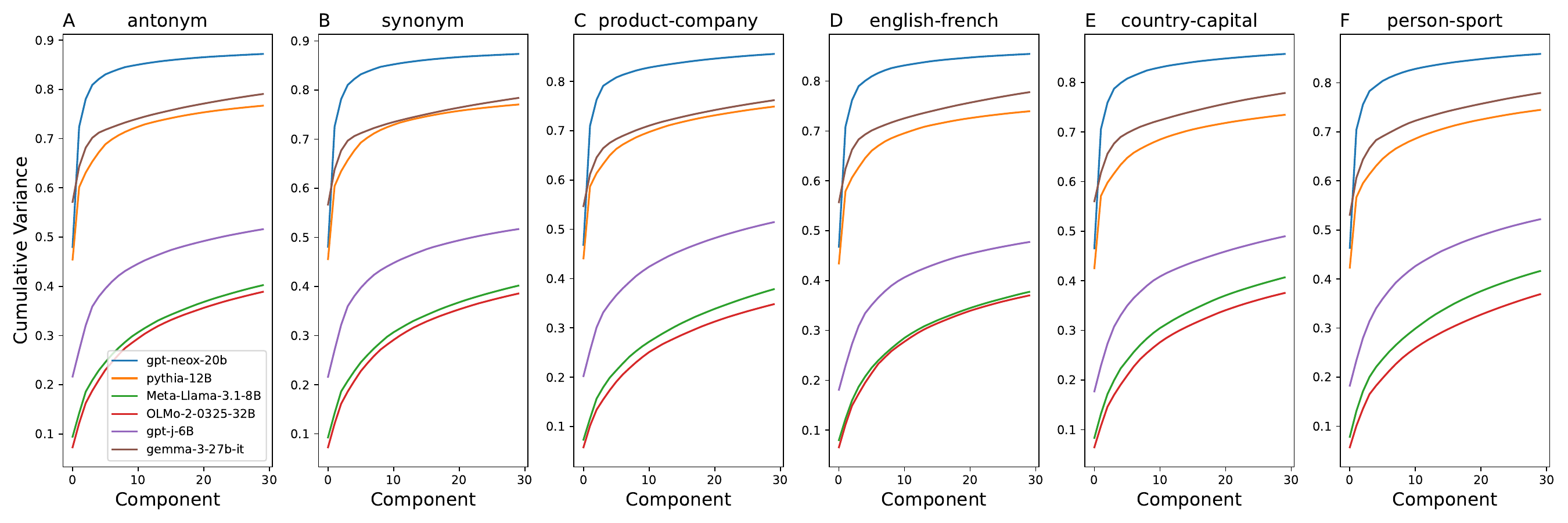}
	\end{center}
	\caption{Explained variance accounted for by $30$ principal components. $x$-axis denotes principal components, and $y$-axis denotes the cumulative explained variance. (A), Explained variance evaluated from LLMs engaging the antonym task. In the panel, 6 models are clarified using different colors (see the inset). (B)-(F), the same as (A), but the task is synonym, country-capital and English-French, country-capital and person-sport, respectively. }\label{fig:2}
\end{figure}

Next, $30$ principal components are evaluated using scikit-learn, an open-source machine learning library\cite{scikit-learn}. Fig. \ref{fig:2} shows the accumulated explained variance for $30$ principal components. As shown in the figure, $40-90\%$ of variability is accounted for by just $30$ principal components, although $d \approx 5000$, suggesting that LLMs can be mapped onto low-dimensional subspace.  

\subsection{ICL Tasks translated as vector algebra problems in subspaces}
Although $30$ principal components can explain the majority of residual streams' variances, it cannot guarantee that LLMs would use them to support ICL. Thus, to further test whether LLMs actively utilize subspaces, we ask if the answer token $\vec{a}$ can be a linear combination of query $\vec{q}$ and separator $\vec{s}$ tokens. When the words are tokenized into multiple tokens, only the last tokens of query and answer words are used in this analysis, which is consistent with the earlier studies \cite{todd2024function,yang2024latentreasoning, meng2023massediting}.  Specifically, for a transformer layer $l$ and a principal component $PC_i$, we project answers, queries and separators in $1000$ examples ($5$ in-context examples in $200$ prompts)  to obtain $\vec{a}^l_i$, $\vec{q}^l_i$, $\vec{s}^l_i$ and regress $\vec{a}^l_i$ into $\vec{q}^l_i$, $\vec{s}^l_i$ (Eq.  \ref{eq_lr2}). 
\begin{equation}\label{eq_lr2}
	\vec{a}^l_i=\alpha \vec{q}^l_i+\beta \vec{s}^l_i+\gamma^l_i, 
\end{equation}
where $\vec{a}^l_i$, $\vec{q}^l_i$, $\vec{s}^l_i$ denote 1-$d$ vectors whose length is $1000$  ($5$ in-context examples in $200$ prompts).

\begin{figure}
	\begin{center}
		\includegraphics[width=1\linewidth]{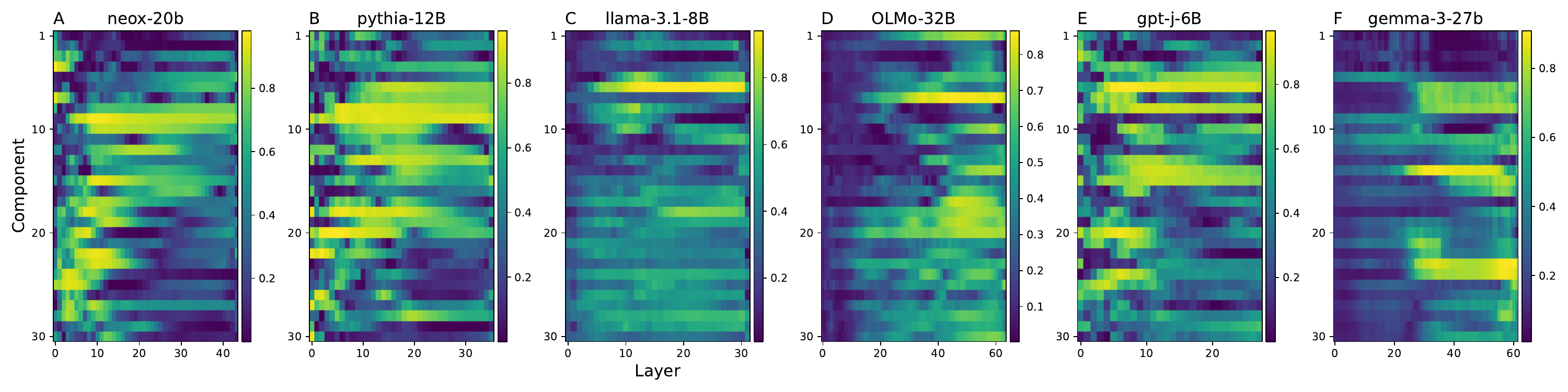}
	\end{center}
	\caption{Quality of linear regression evaluated. $R^2$ is evaluated using LLMs engaging in `antonym' task. $x$-axis denotes the transformer layer, and $y$-axis denotes the principal components. (A)-(F) shows $R^2$ estimated from GPT-j-6B, Meta-Llama-3.1-8B, OLMo-2-$0325$-32B,  Phythia-12B, gemma-3-27b-it and GPT-NEOX-$20$B.}  
	\label{fig:3}
\end{figure}

To evaluate the quality of regression and identify a subspace, where the answer token can be described by the linear combination of query and separator tokens, we use $R^2$. When the regression is perfect, $R^2=1$. By contrast, $R^2$ approaches $-\infty$, when the regression fails. That is, if $R^2$ is close to 1, the principal component can be a basis for a subspace that we aim to identify. Fig. \ref{fig:3} shows $R^2$ collected from all layers of 6 models engaging in the `antonym' task. Across all 6 models, $R^2$ is high along with a few components. That is, answer tokens are well approximated by linear combinations of query and separator tokens (Eq. \ref{eq_lr2}) in a subspace spanned by these components, supporting that LLMs do use subspaces and that ICL tasks can be solved via simple vector operations. We made two more observations. First, $R^2$ is low in early layers and becomes higher in late layers, raising the possibility that the early layers of LLMs gradually transform LLM's activation spaces into subspaces, where ICL can be solved by simple vector operations. Second, the principal components with high $R^2$ appear to be largely consistent across the layers. For instance, $R^2$ is consistently high along component $9$ of GPT-NEOX-$20$B (Fig. \ref{fig:3}A). 

\begin{figure}
	\begin{center}
		\includegraphics[width=1\linewidth]{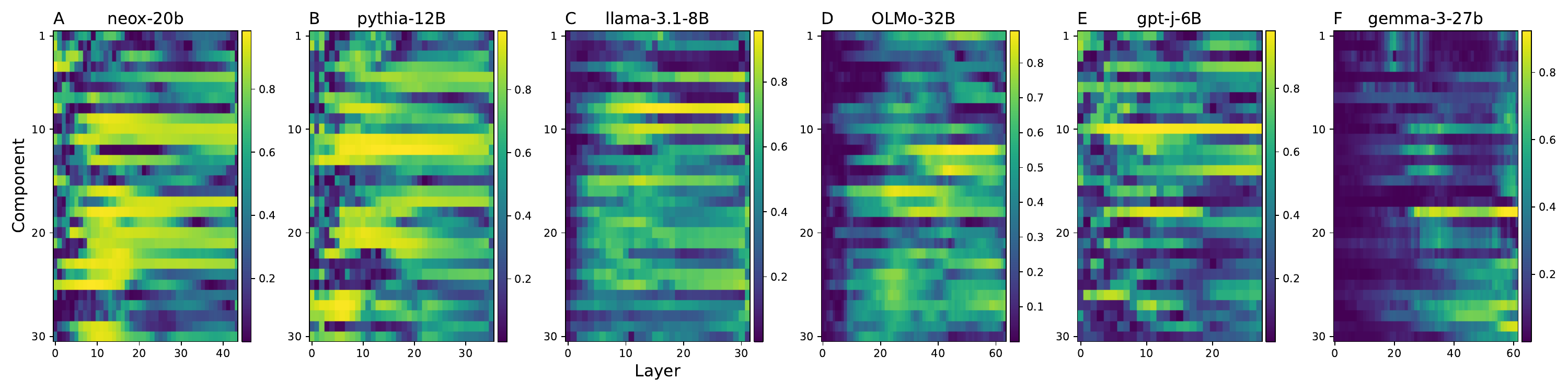}
	\end{center}
	\caption{The same as Fig. \ref{fig:3}, but the task is country-capital.}
	\label{fig:4}
\end{figure}

These observations suggest that the components with high $R^2$ encode the concept `antonym', which is consistent with linear representation hypothesis \cite{park2023the} but further suggests that 1) there are multiple directions associated with concepts and 2) actual computations associated with concepts occur according to these directions. We repeat the same analysis with 5 more tasks and observe the equivalent results with different tasks, `country-capital' (Fig. \ref{fig:4}),  `English-French' (Supplementary Fig. \ref{fig:5}),  `synonym' (Supplementary Fig. \ref{fig:6}), `product-company' (Supplementary Fig. \ref{fig:ns-1}) and 'person-sport' (Supplementary Fig. \ref{fig:ns-2}). 

\begin{table}
	
	\caption{Comparison of $R^2$ with the baselines. We compares the mean values ($V_1$) of top-3  from our original experiments and those ($V_2$) from the control experiments. For each model and task, we display ($V_1$, $V_2$). That is, the first number in the parenthesis denotes the mean value of top 3 $R^2$ from the original experiment, whereas the second number denotes the mean value of top 3 $R^2$ from the control experiment. ANT, SYN, CC, EF, PC, PS stand for antonym, synonym, country-capital, English-French, Product-Campany and Person-Sport, respectively. Neox, Olmo, Py, GPT, Llma and Gamma denote GPT-neox-20B, OLMo-2-$0325$-32B, pythia-12B Llma-3.1-8B and gemma-3-27b-it, respectively.  
	}\label{tab:control}
	\begin{center}

		\begin{tabular}{ccccccc}
			\hline
			Model  &  ANT &  SYN&  CC&  EF&  PC& PS\\\hline
			Neox& (0.97, 0.02)&(0.97, 0.01)&(0.98, 0.01)&(0.97, 0.01)&(0.96, 0.01)&(0.99, 0.02)\\ 
			Olmo& (0.86, 0.01)&(0.8, 0.01)&(0.89, 0.01)&(0.9, 0.01)&(0.85, 0.02)&(0.95, 0.02)\\
			Pyt & (0.96, 0.01)&(0.97, 0.01)&(0.99, 0.02)&(0.97, 0.01)&(0.98, 0.01)&(0.99, 0.01)\\
			GPT& (0.97, 0.01)&(0.95, 0.01)&(0.98, 0.01)&(0.96, 0.01)&(0.96, 0.01)&(0.97, 0.01)\\
			Llma& (0.94, 0.01)&(0.81, 0.01)&(0.96, 0.01)&(0.88, 0.01)&(0.85, 0.01)&(0.97, 0.02)\\
			Gemma& (0.91, 0.01)&(0.91, 0.01)&(0.92, 0.01)&(0.65, 0.01)&(0.89, 0.01)&(0.94, 0.02)\\ \hline
		\end{tabular}
	\end{center}
	
\end{table}

\subsection{Token types in subspaces }
\begin{figure}
	\begin{center}
		\includegraphics[width=0.9\linewidth]{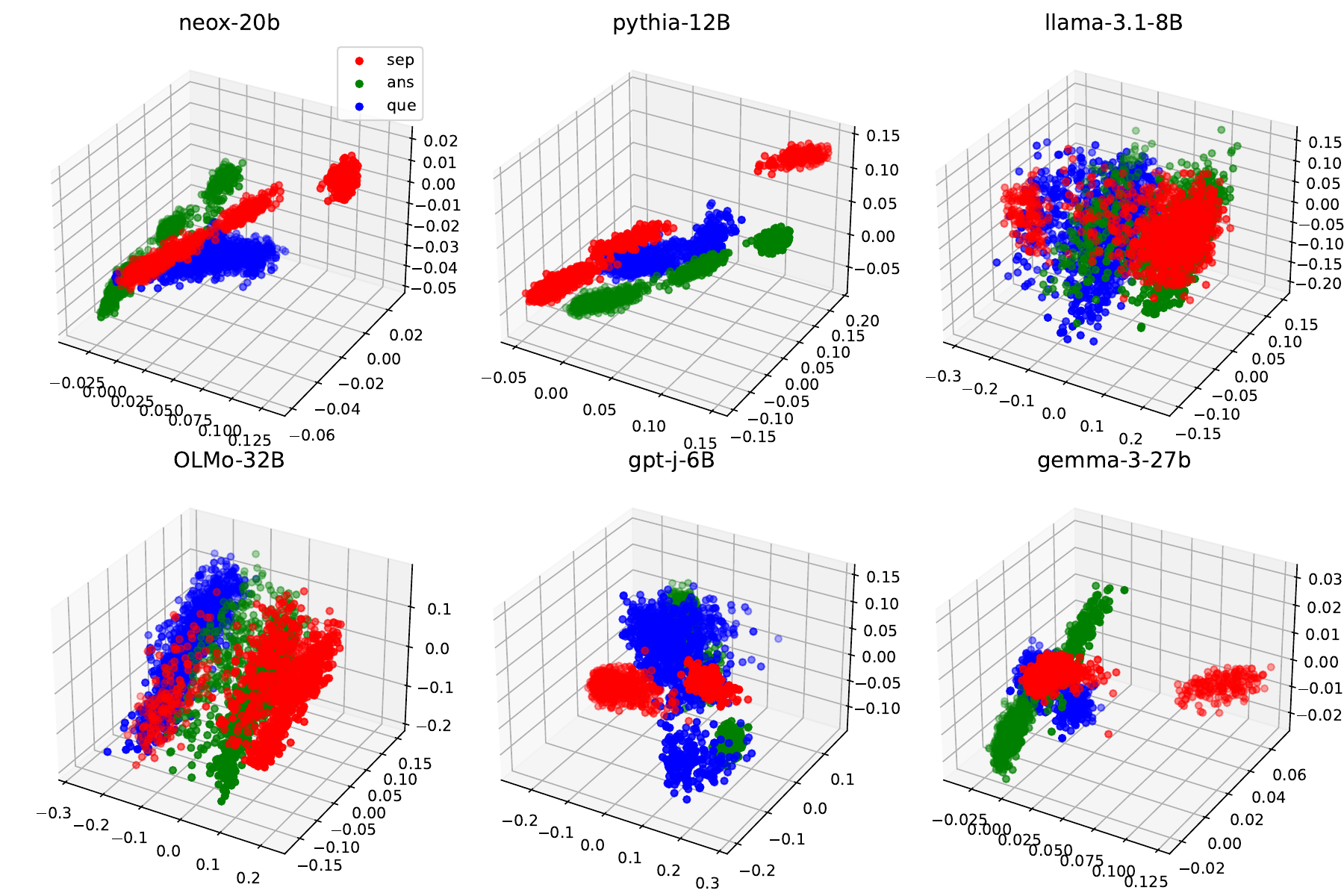}
	\end{center}
	\caption{Query (que), separator (sep) and answer (ans) tokens in subspace spanned by 3 principal components with the highest $R^2$s. Red, green and blue dots represent separator, answer and query tokens, respectively (see the inset). All tokens are collected from $6$ LLMs engaging in the antonym task. The model is specified by the name above each plot.}
	\label{fig:7}
\end{figure}

If LLMs do utilize subspaces and vector operations in it, one can expect that tokens are clustered together depending on types. We test this possibility by visualizing answer, separator and query tokens in $3D$ subspace spanned by the top 3 principal components according to $R^2$. In this analysis, we choose a single layer with the highest $R^2$ from individual models engaging in antonym (Fig. \ref{fig:7}), country-capital (Fig. \ref{fig:8}), English-French (Supplementary Fig. \ref{fig:9}), synonym(Supplementary Fig. \ref{fig:10}), product-company (Supplementary Fig. \ref{fig:ns-3}) and person-sport (Supplementary Fig. \ref{fig:ns-4}). As shown in the figures, answer, separator and query tokens are well clustered with the same token type, and the token types are separated from other types. These results suggest that tokens encode task specific information (i.e., query, separator or answer), supporting that the subspace is associated with LLMs' decision making. 

\subsection{Control experiments}
Finally, to test if $R^2$s are statistically meaningful and higher than random baselines, we conduct control experiments by shuffling $\vec{a}^l_i$. As $\vec{a}^l_i$ is randomized, the linear regression analysis tries to predict answer tokens using incorrect queries and separator tokens, and thus $R^2$ can serve as the baseline of $R^2$ for stochastic noises. To compare $R^2$ with the baselines, 1) we evaluate top 3 $R^2$ with the correct and randomized answer tokens from all layers of models and all $30$ components, and 2) we report the mean values of these top 3 $R^2$ in Table  \ref{tab:control}. As shown in the table, $R^2$s are close to $0$ in the control experiments with randomized answer tokens, whereas $R^2$ are close to 1 in the original experiments with correct answer tokens. This result confirms that $R^2$s, which are evaluated using the correct $\vec{a}^l_i$, are not stochastic artifacts. Additionally, we also replace answer tokens $\vec{a}^l_i$ with  neutral tokens, which are `Q', `A' and line breakers. Then, we replace each answer token with one of the neutral tokens collected from the same input prompt to obtain $R^2$ and find that $R^2$ is close to $0$ in this additional control experiment (Supplementary Table \ref{tab:supp}). 

\subsection{Causal links between subspaces and ICL tasks}
The results above suggest that LLMs can use some subspaces to support ICL, which leads us to assume that perturbation on subspaces would have substantial effects on LLMs' predictions. With this possibility in mind, we test our hypothesis further by perturbing subspaces. Specifically, we edit the last separator tokens $\vec{s}$ in the subspaces spanned by top-3 and bottom-3 components. We choose top-3 components with the highest $R^2$ and bottom-3 components with the lowest $R^2$. Following Alg. \ref{alg1}, we evaluate the perturbed $\vec{s}_{top}$ and $\vec{s}_{bottom}$ for the top-3 components and the bottom-3 components. It should be noted that perturbation $\delta_{bottom}$ and $\delta_{top}$ have the same magnitude. 

Using both $\delta_{top}$ and $\delta_{bottom}$, we compare the subspace, where ICL tasks are solved using linear vector operations, to the subspace, which is weakly associated with ICL tasks. In this experiment, we test 4 models with 3 tasks using $100$ ICL prompts by perturbing the last query $\vec{s}$  in 5 intermediate layers. For the same prompt, we perturb top-3 components and the bottom-3 components independently, and the results are shown in Table \ref{tab:perturb}.

Although we perturb the last separators in 5 layers only, LLMs change their answers almost always when the separator is perturbed via top-3 components. By contrast, when the separators are perturbed via bottom-3 components, LLMs do not change their answers at all (Table \ref{tab:perturb}).These results suggest that the subspace can have causal impacts on LLMs’ decision-making.

\begin{table}
	
	\caption{This table shows how often LLMs change their answers when the last token is perturbed via top-3 and bottom-3 components. For each case, the first number in the parenthesis shows the ratio of the trials, in which LLMs change their answers when the perturbation is delivered via top-3 components. The second number shows the ratio of the trials, in which LLMs change their answers when perturbation is delivered via bottom-3 components. PS, CC and PC stand for Person-Sport, Country-Capital and Product-Company, respectively.}\label{tab:perturb}
	\begin{center}
		\begin{tabular}{cccc}
			\hline
			Model&  PS&  CC& PC\\ \hline
			pythia-12B &  (1.0, 0.1)&  (1.0, 0.25)& (1.0, 0.16)\\
			gpt-neox-20b&  (1.0, 0.09)&  (0.96, 0.08)& (1.0, 0.18)\\
			gpt-j-6B&  (0.98, 0.12)&  (0.99, 0.01)& (1.0, 0.08)\\
			gemma-3-27b-it&  (0.7, 0.02)&   (0.72, 0.0)& (0.09, 0.06)\\ \hline
		\end{tabular}
		
	\end{center}
\end{table}

\begin{algorithm}
	\caption{Algorithm used to create perturbation. We perturb layers $25-29$ of gemma-3-27b-it and layers $11-15$ of pythia-12B, gpt-neox-20b and gpt-j-6B. }\label{alg1}
	\begin{algorithmic}
		\State 1. The last token token $\vec{s}$ is normalized $\vec{s} \leftarrow N(\vec{s})$ and centered $\vec{s} \leftarrow S(\vec{s})$
		
		\State 2. Project $\vec{s}$ to top-3 components $V_0$, $V_1$ and $V_2$ with highest $R^2$ and obtain coefficients $c_0$, $c_1$ and $c_2$.
		\State 3. Estimate perturbation $\delta_{top}=c_0V_0+c_1V_1+c_2V_2$.
		\State 4. Obtain perturbation for top-3 components, $\vec{s}_{top} \leftarrow \vec{s}+\delta_{top}$.
		\State 5. Project the last separator token $\vec{s}$ to bottom-3 components $B_0$, $B_1$ and $B_2$ with lowest $R^2$ and obtain coefficients $b_0$, $b_1$ and $b_2$. 
		\State 6. Estimate perturbation $\delta_{bottom}=sign(b_0)|c_0|B_0+sign(b_1)|c_1|B_1+sign(b_2)|c_2|B_2$.
		\State 7. Obtain Perturbation for bottom-3 components,  $\vec{s}_{bottom} \leftarrow \vec{s}+\delta_{bottom}$.
		\State 8. Project $\vec{s}_i$ back into token space with $\vec{s} \leftarrow S^{-1}(\vec{s}_i)$ and $\vec{s} \leftarrow N^{-1}(\vec{s}_i)$, where $i$ is either top or bottom.
		
	\end{algorithmic}
\end{algorithm}

We additionally test the correlations between subspaces and LLMs' decisions and report the results in \ref{a1}.

\section{Discussion}
Our analyses suggest that tokens in ICL prompts can be encoded in subspaces and that their relations are explained by the linear operations. Below, we discuss the implications of our analyses for ICL capabilities of LLMs, links to the earlier works and limitations. 

\subsection{Subspace and ICL}
The residual stream of a LLM, which stores activations from all of its layers, allows its last layer to effective access all subspaces $\{\Omega_i \}$. This raises the possibility that LLMs can encode and retrieve the relevant information regarding entities in the input prompt in parallel. For example, the Space Needle in an input prompt may activate multiple subspaces associated with the Space Needle. Two of them can be the location subspace $\Omega_1$ to recall, which encodes that the Space Needle is located in Seattle and the facility function subspace  $\Omega_2$, which encodes that the Space Needle is an observatory. 

It seems natural to assume that the degree of activation strengths of subspaces $\Omega_1$ and $\Omega_2$ would follow the distribution of the training set. If the ‘Space Needle’ and ‘Seattle’ appear together 40\% of the time in the training set, and the ‘Space Needle’ and the ‘observatory’ appear together 20\% of the time, $\Omega_1$ is activated two times more strongly than $\Omega_2$. However, if the ‘Space Needle’ is paired with 'Seattle' in the input prompt, $\Omega_1$ would be more strongly activated. 

With this possibility in mind, we propose that the biased competition among subspaces can underlie ICL capabilities of LLMs. The last query in the prompt can activate many subspaces $\{ \Omega^{q}_i\}$, each of which encodes information associated with the query. Each example (i.e., pair of a query and an answer) can activate its own set of subspaces $\{ \Omega^{e_j}_i\}$ . If a common subspace  $\Omega^{q _i\cap e_i}_i$ is jointly activated by the query and the example, this subspace will become strongly activated and likely become a winner which determines the LLMs' answer to the final query. When more examples are given, the common subspaces $\Omega^{q _i\cap \{e_i\} }_i$ are more likely to exist. Using $\Omega^{q _i\cap \{e_i\} }_i$, the LLM can find more accurate answer to the last query.

\begin{figure}
	\begin{center}
		\includegraphics[width=0.9\linewidth]{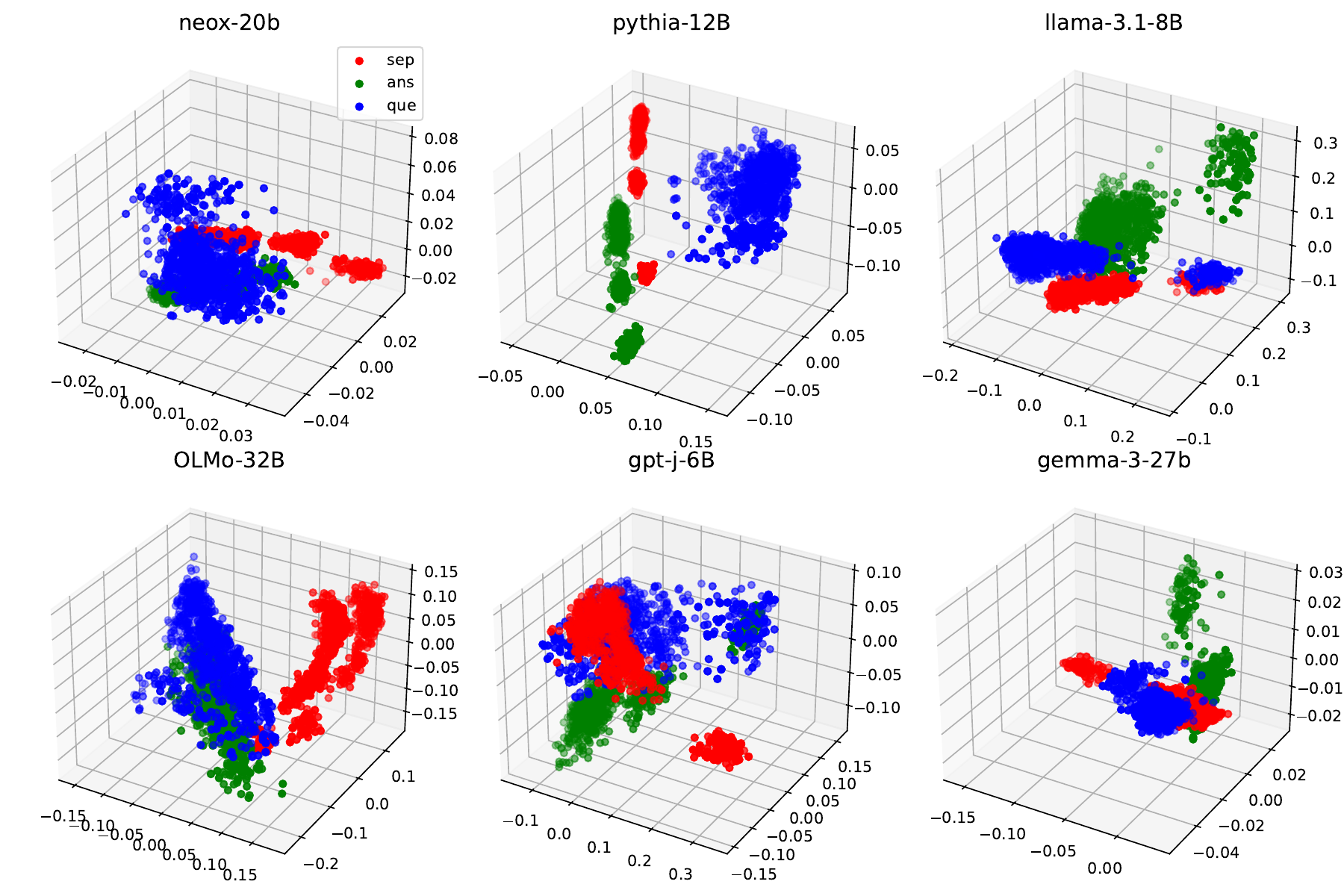}
	\end{center}
	\caption{The same as Fig. \ref{fig:7}, but the task is country-capital. }
	\label{fig:8}
\end{figure}

\subsection {Related Work}
A line of studies suggests that high level concepts are encoded in DNNs \cite{park2023the, jiang2024on, Alain2016,TCAV}. Platonic representation hypothesis \cite{pmlr-v235-huh24a} posits that neural networks can converge to a `shared statistical model of reality in their representation space’, even when they have different architectures and are trained independently. Linear representation hypothesis \cite{pmlr-v235-park24c} further proposes that counterfactual concepts are encoded linearly in LLM activation space. However, it remains unclear how concept directions emerge in LLM activation space and whether concepts could be encoded via directions of single vectors \cite{zhao2025beyond, engels2025not}. 

Our analysis of ICL provides valuable insights into underlying mechanisms by which LLMs generate linear representations and their significance in their learning and generalization capabilities. After LLMs are pretrained to predict next tokens, residual streams may encode potential answers in parallel and thus can be decomposed into a set of components corresponding to potential answers. In our analysis, tokens in the input prompts can be mapped onto the same subspaces (Figs. \ref{fig:7} and \ref{fig:8}), and linear operations can solve the desired tasks (Figs. \ref{fig:3} and \ref{fig:4}). Although our analysis focuses on ICL, it is based on generic properties of LLMs and thus, can be generalized to other language tasks. If this generalization holds, our analysis raises the possibility that LLMs map information (i.e., tokens) existing in input prompts onto data points in subspaces, whose geometry is directly associated with desired tasks. In this view, a linear representation of a concept emerges naturally in a subset of subspaces, which are functionally crucial to desired tasks. Additionally, we note that LLMs can create a large number of subspaces and track them all in parallel via residual connections, and multi-head attention can create independent subspaces for the same inputs. Together with our analysis, we propose that these subspaces can perform distinct functions and that as a result LLMs can readily implement a wide range of functions after pretraining. 

Below, we review two lines of studies on ICL and the potential links between our work and them

First,  Xie et al. \cite{icl-bayesian} proposed that ICL can be considered as implicit Bayesian inference. Specifically, they noted that LLMs may learn the distribution of latent concepts $\Theta$, and for a $\theta \in \Theta$, LLMs can generate output tokens using latent concept space: $p(O_1, O_2, \ldots, O_T)=p(O_1, O_2, \ldots, O_T|\theta)p(\theta)$. In principle, Bayesian inference rule allows us to predict $p(\theta| O_1, O_2, \ldots, O_T)$ from $p(O_1, O_2, \ldots, O_T|\theta)p(\theta)$. They provided theoretical arguments how to use derive a formal framework on ICL from Bayesian rule. This study, however, does not present how LLMs can recover a latent concept $\theta$ from ICL examples. Our analyses suggest that LLMs use subspaces to encode concepts. Conversely, the collection of subspaces can map onto latent concept space $\Theta$.  

Second, another theoretical view is based on dual gradient updates \cite{dual-gradient, dai-etal-2023-gpt}. Dai et al. \cite{dai-etal-2023-gpt} proposed evidence supporting strong `ICL-GD (gradient descent) correspondence' in pretrained LLMs by probing the outputs and updates of attention heads. Specifically, they found that the updates and outputs of attention heads are well aligned between ICL and fine-tuning. However, this evidence for ICL-GD correspondence was questioned by a more recent study \cite{dual-gradient-skeptic}. Although it is still unclear if dual gradient descent underlies ICL, it may explain how LLM generate subspaces, where simple vector operations can solve ICL tasks. 

We also discuss the limitations of our study in \ref{a2}

\bibliography{ref}

\bibliographystyle{unsrt}


\appendix

\setcounter{figure}{0}
\renewcommand{\figurename}{Supplementary Figure}

\setcounter{table}{0}
\renewcommand{\tablename}{Supplementary Table}

\section{Appendix}
\subsection{Correlations between subspaces and LLMs' decisions}\label{a1}

If subspaces are correlated with LLMs' decision-making, one may expect the tokens' projections to be `significantly' different depending on the correctness of LLMs' answers. Thus, we compare  all three tokens  ($a^l$, $s^l$ and $q^l$) depending on the correctness of LLMs' answers. Specifically, we choose top-1 component in each layer $l$ with highest $R^2$, project the last set of query, separator and answer tokens to the component and compare them under two conditions, when LLMs make correct predictions and when they make incorrect ones. It should be noted that the last answer token is obtained by allowing LLMs to generate new tokens and that $t$-test is used to evaluate whether $a^l$, $s^l$ and $q^l$ are significantly different from each other depending on the correctness of LLMs' answers. 

\begin{figure}
	\begin{center}
		\includegraphics[width=1\linewidth]{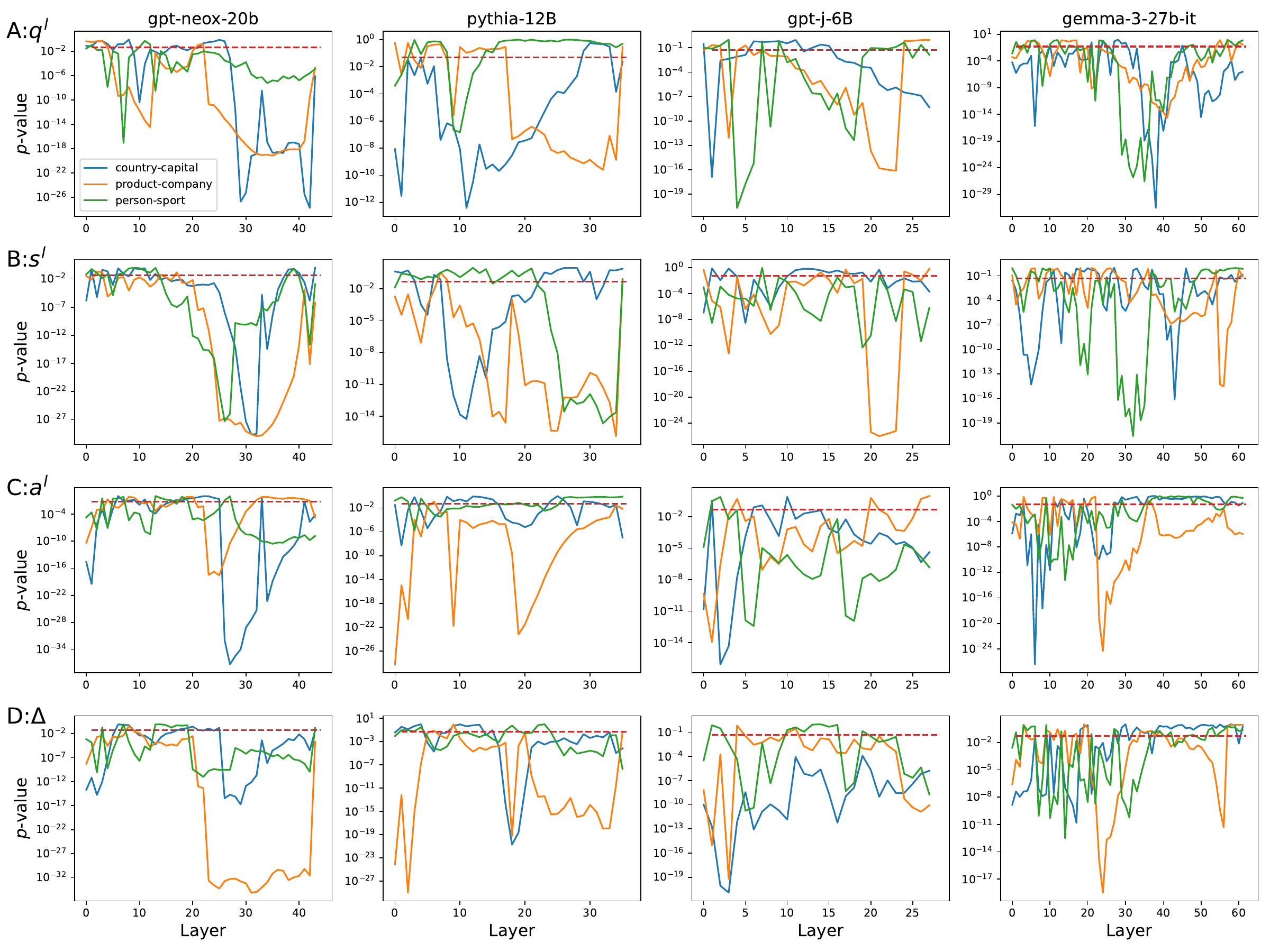}
	\end{center}
	\caption{Comparison of query $q^ l$ in Panel (A), separator $s^l$ in Panel (B), answer $a^l$ in Panel (C),  and $\Delta$ in Panel (D) when LLMs make the correct predictions and when they make incorrect ones. In this experiment, we use $t$-test and report $p$-values. The blue, orange and green lines denote country-capital task, product-company task and person-sport task, respectively.  }
	\label{fig:corr}
\end{figure}

In this experiment, we test 4 models (gpt-neox-20b, pythia-12B, gpt-j-6B and gemma-3-27b-it) and 3 tasks (country-capital, product-company and person-sport) using $400$ ICL prompts, different from $200$ prompts used to evaluate $30$ principal components. Fig. \ref{fig:corr} shows the $p$-value of the t-test on 3 models engaging in 3 tasks. $p$-value presents the probability that the two distributions are drawn from the same distribution. The two distributions are `significantly’ different when $p\leq0.05$. 

As shown in the figure, the $p$-values are below $0.05$ in multiple layers, suggesting that the projections of all three token types along with the top-1 component are significantly different between two conditions, when LLMs make correct predictions and when they make incorrect predictions. Conversely, the computations occurring along with the top-1 component may be correlated with LLMs’ answers. Further, we also compare the error $\Delta=a^l_{j=last}-\alpha q^l-\beta s^l-\gamma^l$ depending on the accuracy of LLMs and find the same trend that $\Delta$ is also significantly (t-test, $p<0.05$) different from one another in multiple layers (Fig. \ref{fig:corr}D). 

We note that $p$-values vary depending on the tasks and the models, but they are consistently below $0.05$ in most of the intermediate and late layers ($layer\geq 20)$, suggesting that the intermediate layers can use linear algebra to infer proper predictions. Interestingly,  $p$-value become closer to $0.05$ when a layer is close to output (i.e., final) layer. Currently, it remains unclear why the late layers close to output layers become less sensitive to the correctness of LLMs' predictions. A potential reason is that the final layers may integrate evidence across examples rather than processing single examples. If this is the case, tokens (and thus, $\Delta$) will encode information from all examples and become less sensitive to vector operations among tokens in a single example. 

Although this analysis is based on a single component, our results raise the possibility that the subspace could be correlated with LLMs’ decision-making.  

\subsection{Limitations}\label{a2}

We acknowledge the following limitations of our study. First, in Section 2, we analyze the transformers’ functional modules (self-attention and FFNs) under an ideal condition, which does not generally hold. We use this abstract model to gain insights into the mechanisms by which the transformers accumulate evidence from examples in ICL prompts. Second, we use ICL tasks as proxy models of language tasks. We plan to extend our approach to more general language tasks but assume that it will be challenging to identify subspaces in general tasks because tokens interact with one another in complex ways. 

\begin{figure}[h]
	\begin{center}
		\includegraphics[width=1\linewidth]{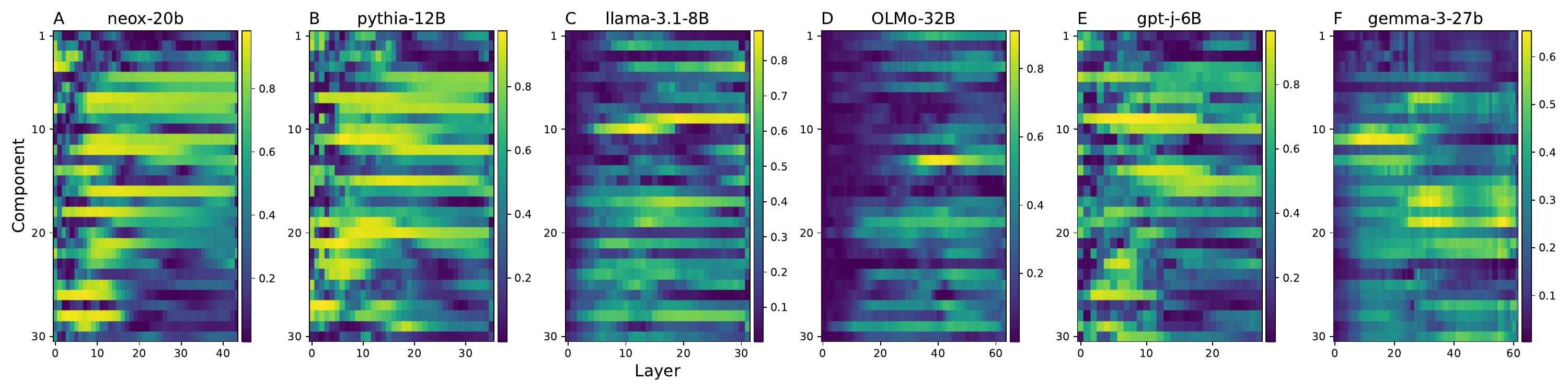}
	\end{center}
	\caption{The same as Fig. \ref{fig:3}, but the task is the English-French.}
	\label{fig:5}
\end{figure}

\begin{figure}[h]
	\begin{center}
		\includegraphics[width=1\linewidth]{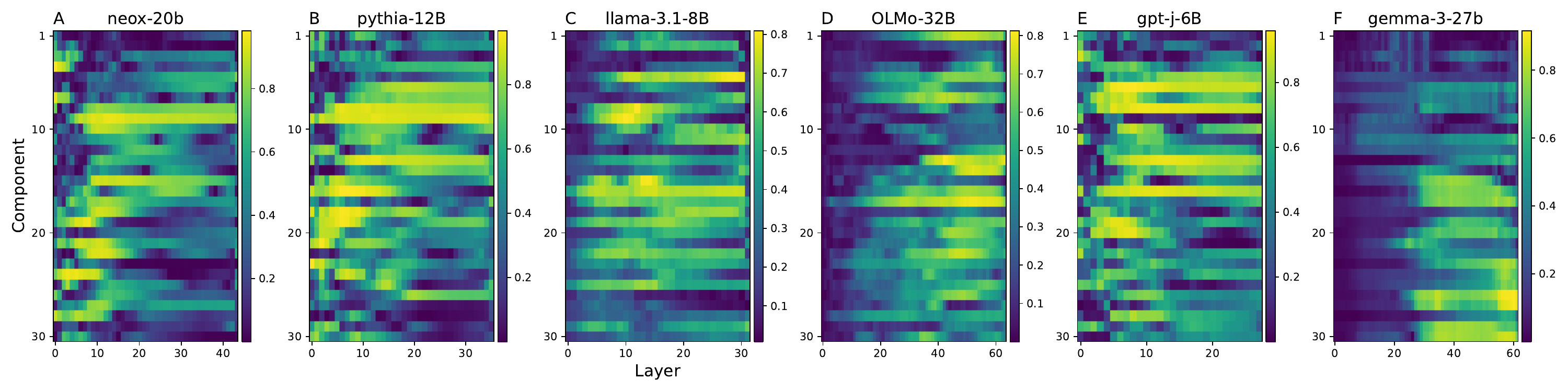}
	\end{center}
	\caption{The same as Fig. \ref{fig:3}, but the task is synonym.}
	\label{fig:6}
\end{figure}
\begin{figure}[h]
	\begin{center}
		\includegraphics[width=1\linewidth]{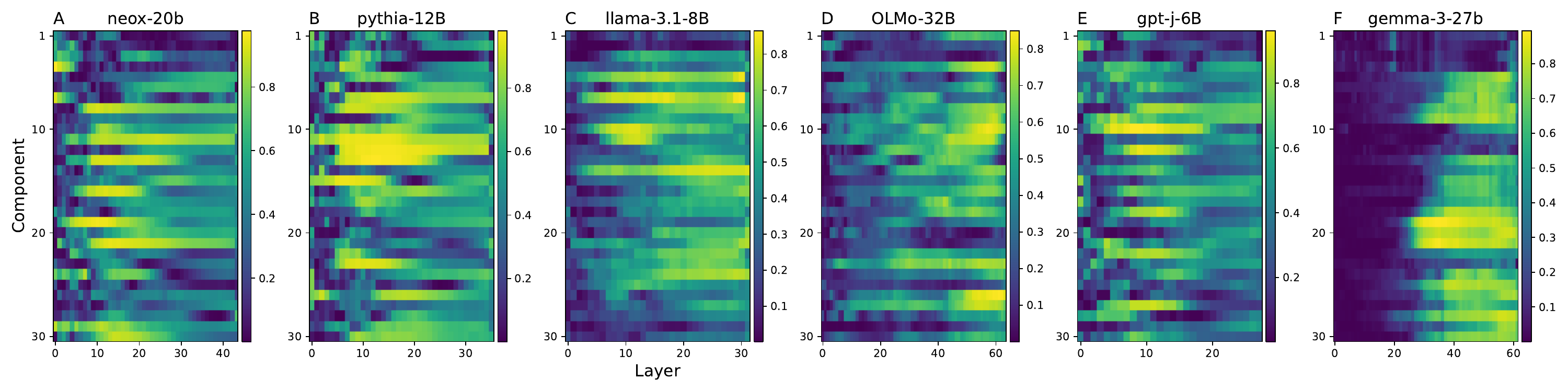}
	\end{center}
	\caption{The same as Fig. \ref{fig:3}, but the task is Product-Company.}
	\label{fig:ns-1}
\end{figure}

\begin{figure}[h]
	\begin{center}
		\includegraphics[width=1\linewidth]{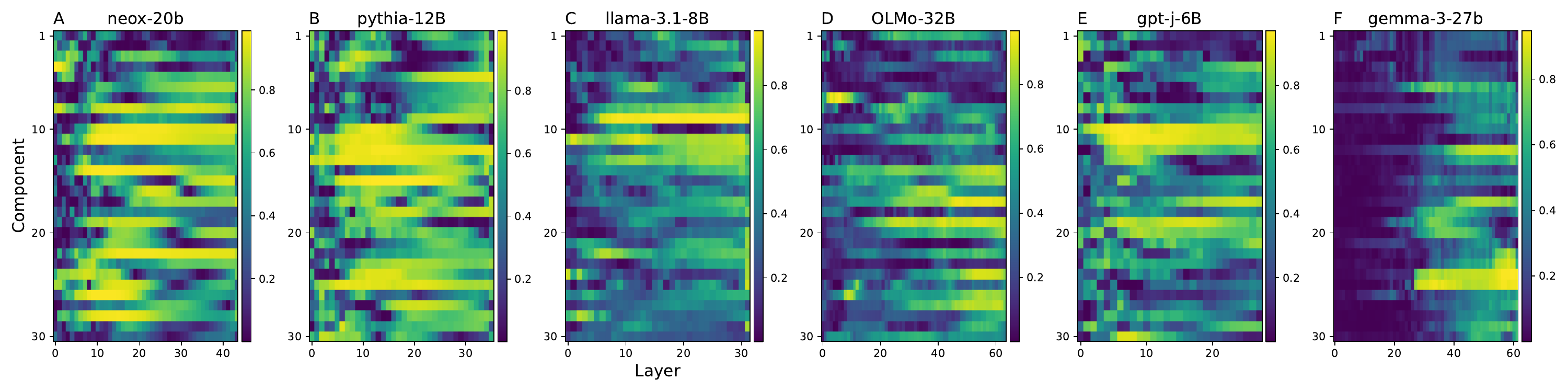}
	\end{center}
	\caption{The same as Fig. \ref{fig:3}, but the task is Person-Sport.}
	\label{fig:ns-2}
\end{figure}

\begin{figure}[h]
	\begin{center}
		\includegraphics[width=1\linewidth]{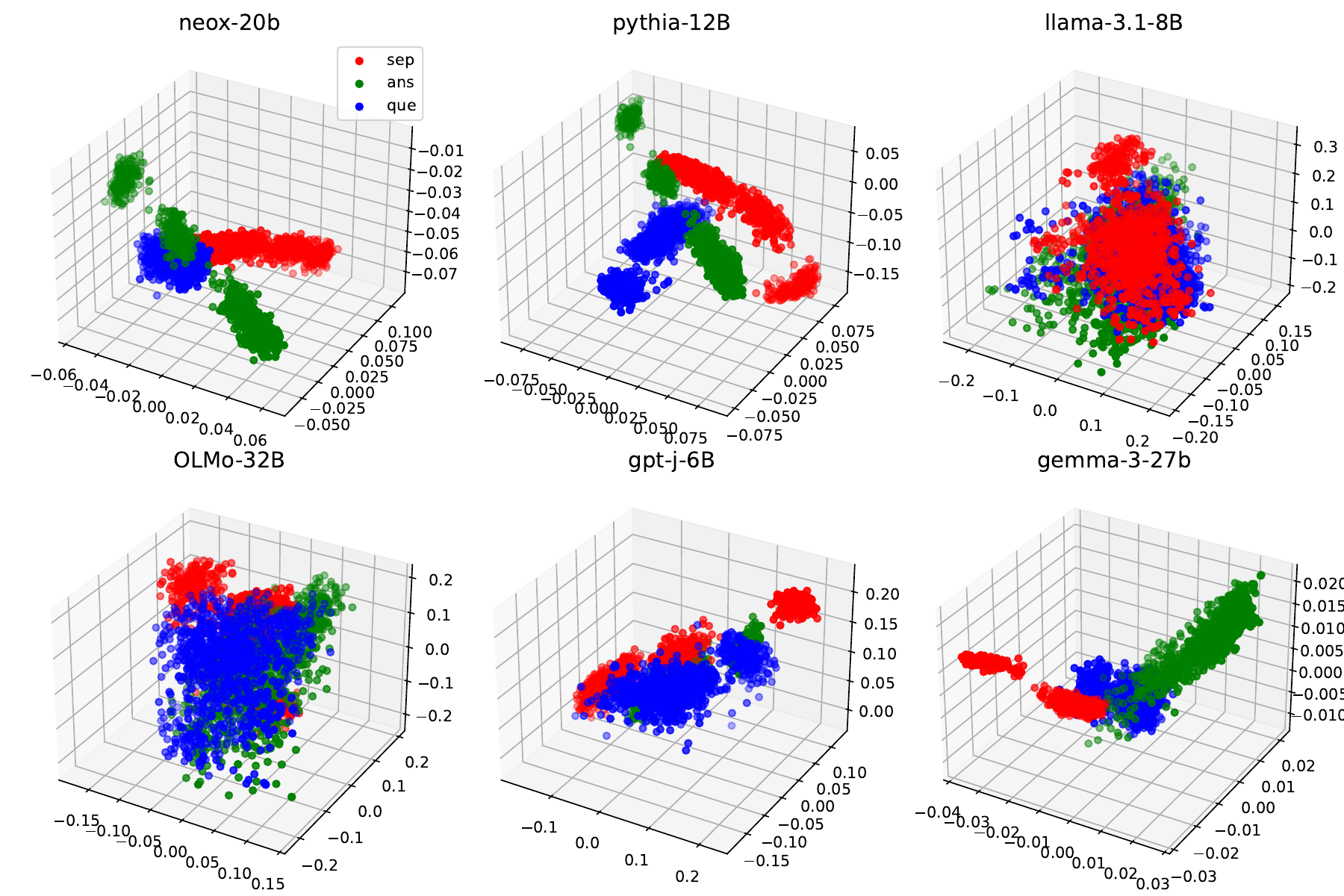}
	\end{center}
	\caption{The same as Fig. \ref{fig:7}, but the task is English-French.}
	\label{fig:9}
\end{figure}

\begin{figure}[h]
	\begin{center}
		\includegraphics[width=1\linewidth]{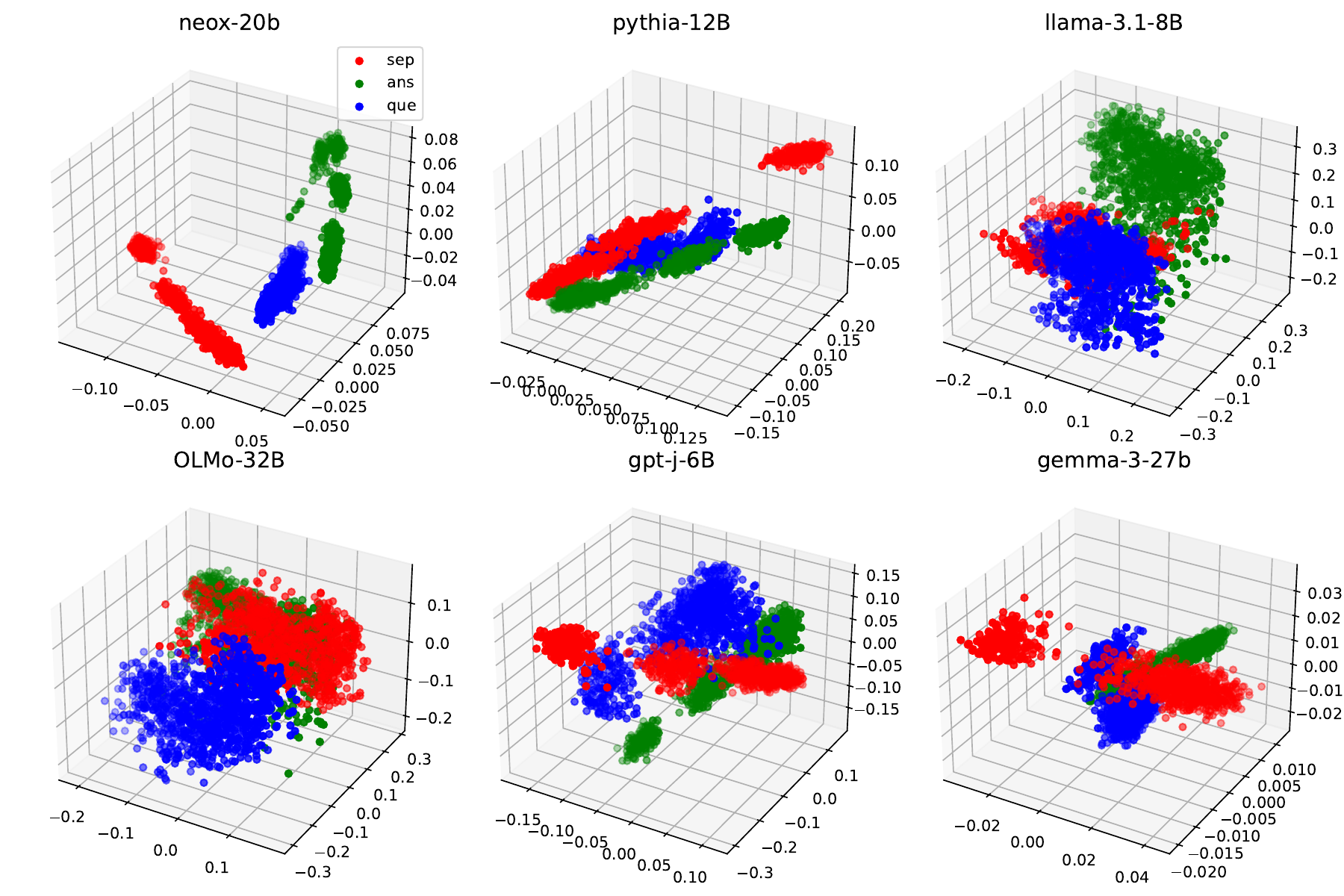}
	\end{center}
	\caption{The same as Fig. \ref{fig:7}, but the task is synonym.}
	\label{fig:10}
\end{figure}

\begin{figure}[h]
	\begin{center}
		\includegraphics[width=1\linewidth]{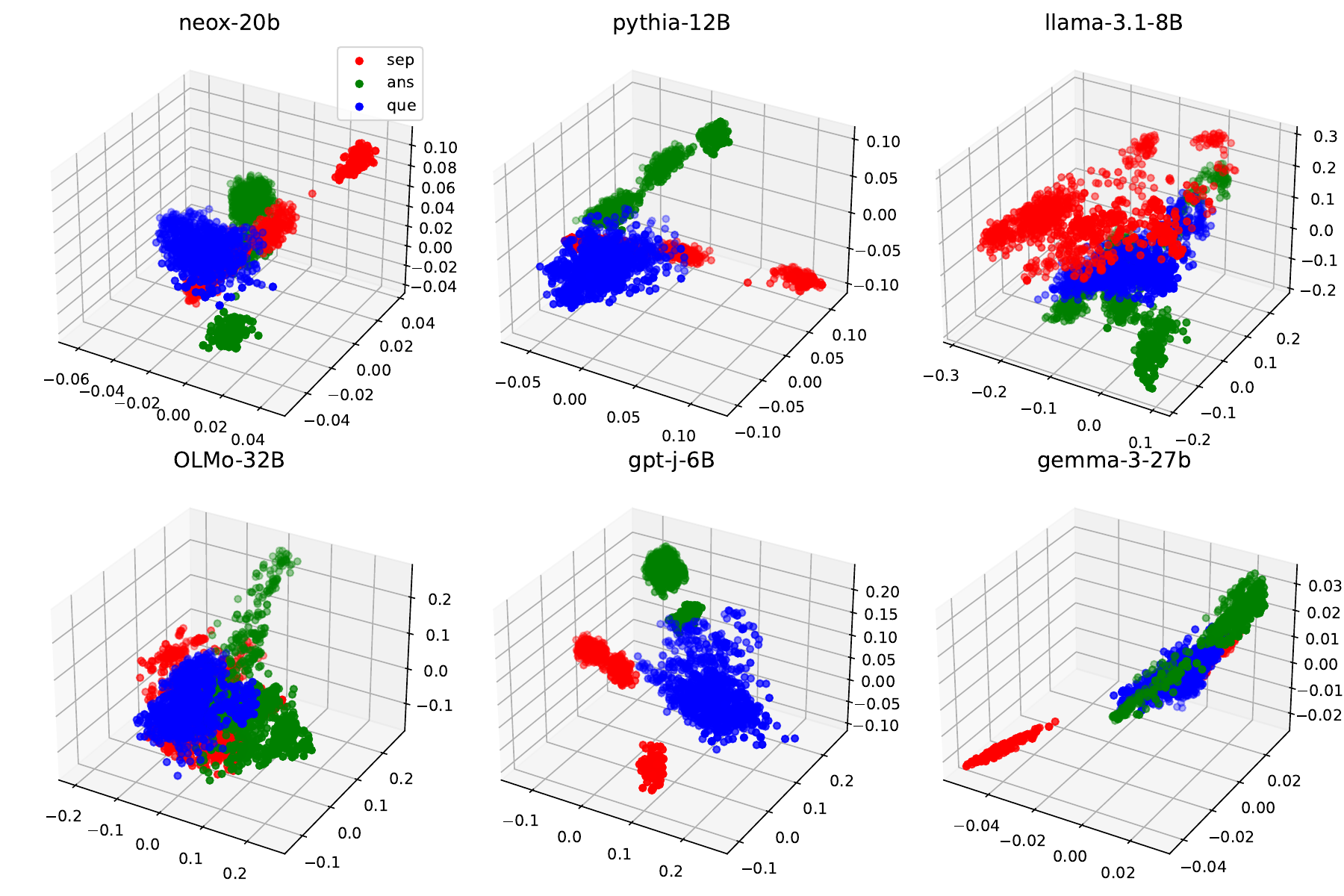}
	\end{center}
	\caption{The same as Fig. Fig. \ref{fig:7}, but the task is product-company.}
	\label{fig:ns-3}
\end{figure}

\begin{figure}[h]
	\begin{center}
		\includegraphics[width=1\linewidth]{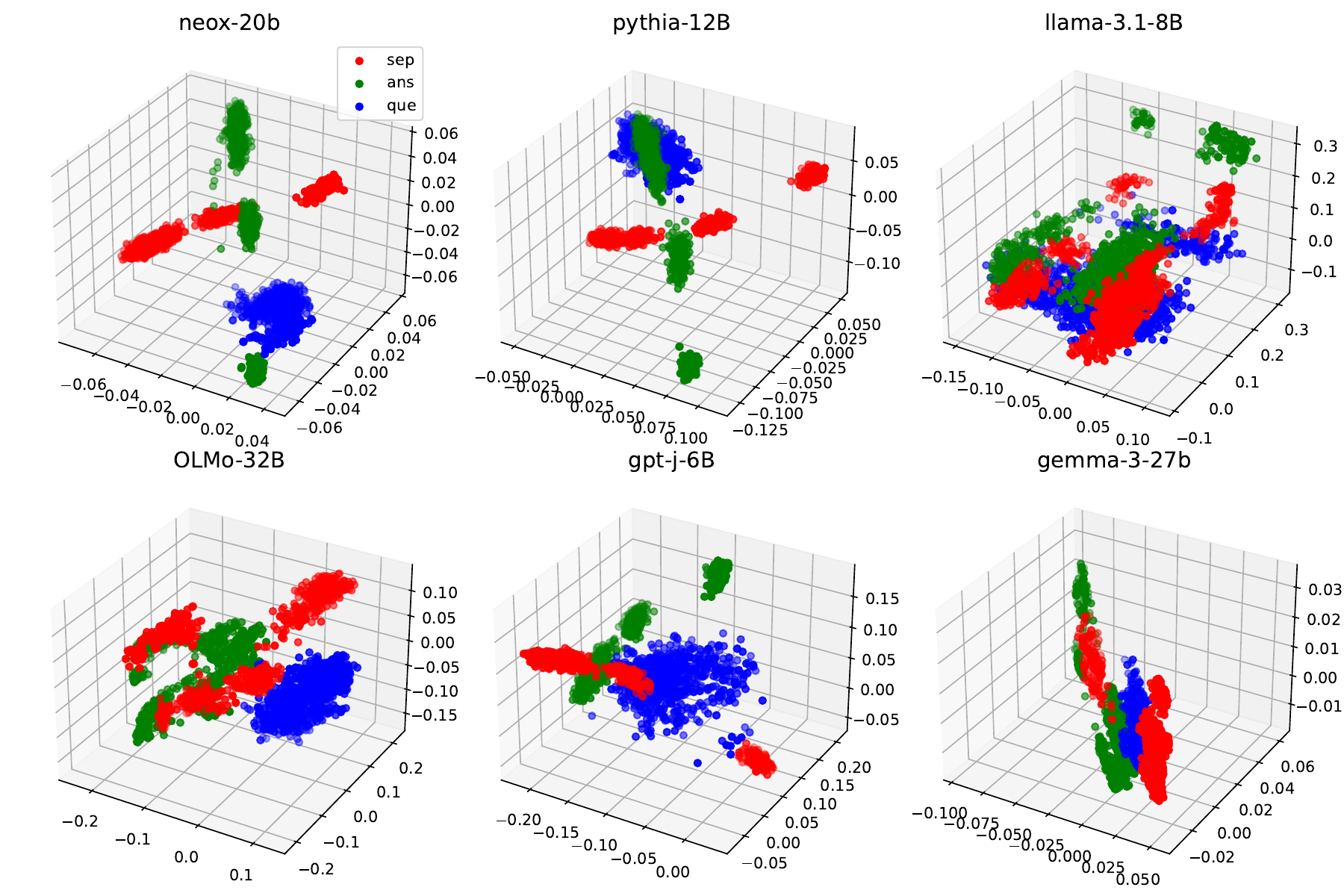}
	\end{center}
	\caption{The same as Fig. Fig. \ref{fig:7}, but the task is person-sport.}
	\label{fig:ns-4}
\end{figure}

\begin{table}
	
	\caption{We compare the mean values ($V_1$) of top-3 $R^2$ from our original experiments and those ($V_2$) from the control experiments. For each model and task, we display ($V_1$,$V_2$ ). That is, the first number in the parenthesis denotes the mean value of top-3  from the original experiments, whereas the second number denotes those from the control experiments. ANT, SYN, CC, EF, PC, PS stand for antonym, synonym, country-capital, English-French, Product-Campany and Person-Sport, respectively.Neox, Olmo, Pyt, GPT, Llama and Gamma denote GPT-neox-20B, OLMo-2-$0325$-32B, pythia-12B Llma-3.1-8B and gemma-3-27b-it, respectively.}\label{tab:supp}
	\begin{center}
		\begin{tabular}{ccccccc}
			\hline
			Model  &  ANT &  SYN&  CC&  EF&  PC& PS\\ \hline
			Neox&(0.97, 0.04)&(0.97, 0.04)&(0.98, 0.05)&(0.97, 0.04)&(0.96, 0.08)&(0.99, 0.06)\\
			Olmo&(0.86, 0.05)&(0.8, 0.03)&(0.89, 0.02)&(0.9, 0.08)&(0.85, 0.02)&(0.95, 0.02)\\
			Pyt &(0.96, 0.03)&(0.97, 0.04)&(0.99, 0.03)&(0.97, 0.02)&(0.98, 0.09)&(0.99, 0.02)\\
			GPT&(0.97, 0.06)&(0.95, 0.04)&(0.98, 0.03)&(0.96, 0.03)&(0.96, 0.09)&(0.97, 0.05)\\
			Llama&(0.94, 0.06)&(0.81, 0.04)&(0.96, 0.02)&(0.88, 0.03)&(0.85, 0.05)&(0.97, 0.03)\\
			Genna&(0.91, 0.03)&(0.91, 0.03)&(0.92, 0.02)&(0.65, 0.02)&(0.89, 0.04)&(0.94, 0.02)\\ \hline
		\end{tabular}
		
	\end{center}
\end{table}

\end{document}